\documentclass[letterpaper]{article} 
\usepackage{aaai2026}  
\usepackage{times}  
\usepackage{helvet}  
\usepackage{courier}  
\usepackage[hyphens]{url}  
\usepackage{graphicx} 
\usepackage{natbib}  
\usepackage{caption} 
\usepackage{algorithm}
\usepackage{algorithmic}
\usepackage{newfloat}
\usepackage{listings}
\DeclareCaptionStyle{ruled}{labelfont=normalfont,labelsep=colon,strut=off} 
\floatstyle{ruled}
\newfloat{listing}{tb}{lst}{}
\floatname{listing}{Listing}
\usepackage{xspace}
\usepackage{xcolor}
\usepackage{multirow}
\usepackage{booktabs}
\usepackage{amsmath}
\usepackage{amssymb}
\usepackage{subcaption}
\usepackage{makecell}
\usepackage{enumitem}
\setlist[itemize]{leftmargin=2.5mm}

\title{\textsc{Atom}: Efficient On-Device Video-Language Pipelines Through Modular Reuse}
\author{
    Kunjal Panchal\textsuperscript{\rm 1}, 
    Saayan Mitra \textsuperscript{\rm 2},
    Somdeb Sarkhel \textsuperscript{\rm 2},
    Haoliang Wang \textsuperscript{\rm 2},
    Ishita Dasgupta \textsuperscript{\rm 2},
    Gang Wu \textsuperscript{\rm 2},
    Hui Guan \textsuperscript{\rm 1}
}
\affiliations{
    \textsuperscript{\rm 1}University of Massachusetts, Amherst
    \textsuperscript{\rm 2}Adobe Research
}

\usepackage{bibentry}

\newcommand{\projectname}[0]{\textsc{Atom}\xspace}

\begin{document}

\maketitle

\begin{abstract}
Recent advances in video-language models have enabled powerful applications like video retrieval, captioning, and assembly. 
However, executing such multi-stage pipelines efficiently on mobile devices remains challenging due to redundant model loads and fragmented execution. 
We introduce \projectname, an on-device system that restructures video-language pipelines for fast and efficient execution. 
\projectname decomposes a billion-parameter model into reusable modules, such as the visual encoder and language decoder, and reuses them across subtasks like captioning, reasoning, and indexing. 
This reuse-centric design eliminates repeated model loading and enables parallel execution, reducing end-to-end latency without sacrificing performance. 
On commodity smartphones, \projectname achieves 27--33\% faster execution compared to non-reuse baselines, with only marginal performance drop ($\leq$2.3 Recall@1 in retrieval, $\leq$1.5 CIDEr in captioning). 
These results position \projectname as a practical, scalable approach for efficient video-language understanding on edge devices.
\end{abstract}


\section{Introduction}
\label{sec:introduction}
Video-language model (VLM) pipelines~\cite{jo2024vvs, li2025tgllava, long2025retrievalvlm} are rapidly becoming the engine behind emerging mobile applications such as on-device video retrieval~\cite{tahboub2015contentbasedvideoretrieval, zhang2021personalizedvideoretrievalmobiledevices} and video assembly~\cite{yang2023shotretrievelandassembly}, where users locate or stitch together video clips through natural-language prompts.
Such applications rely on multi-stage pipelines that typically involve three core subtasks: video encoding, caption generation, and downstream reasoning or indexing.
Running these pipelines locally is highly desirable because it preserves privacy and does not require a cloud server infrastructure. 
\begin{figure}[!h]
    \centering
    \includegraphics[width=\linewidth]{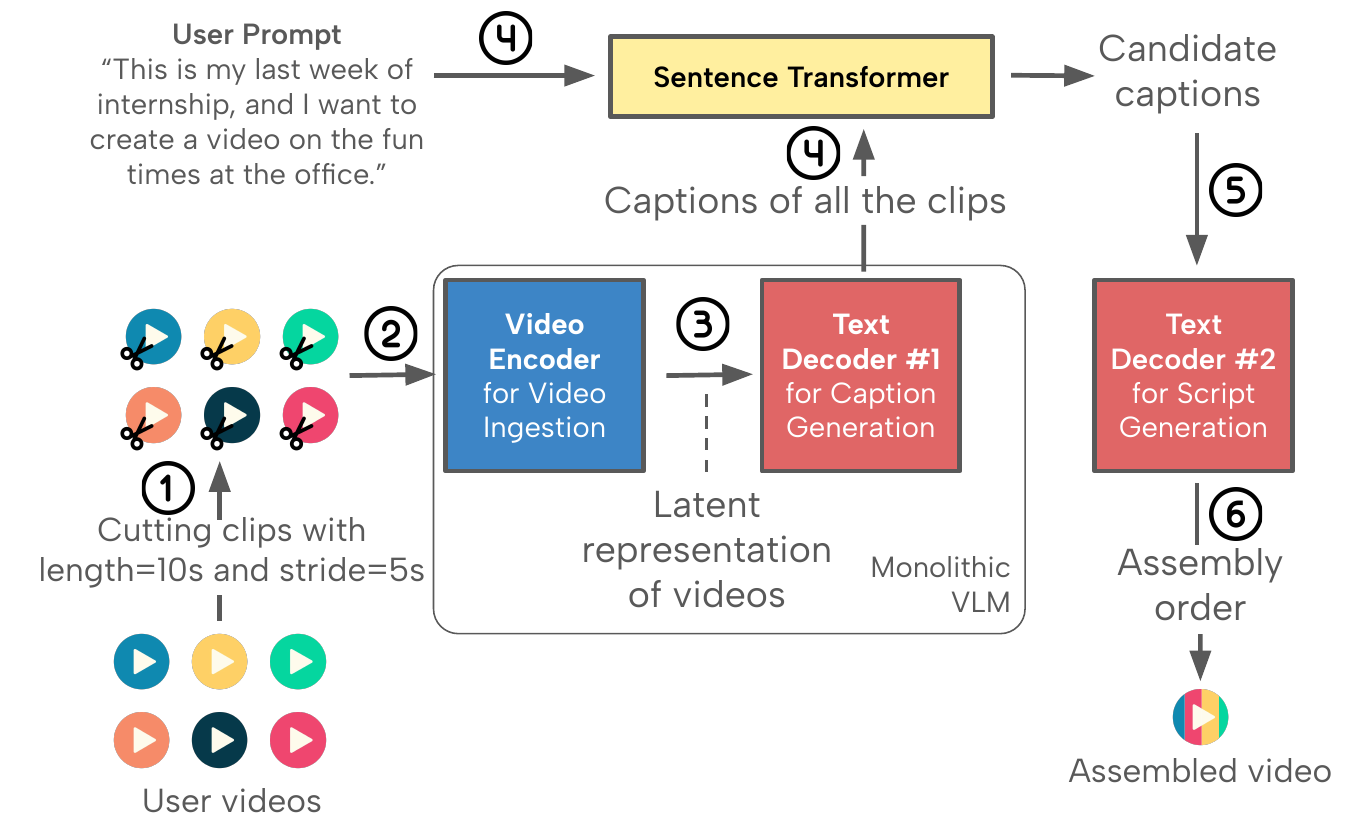}
    \caption{An end-to-end on-device sequential pipeline for video assembly. Given (i) user videos and (ii) a user prompt, the pipeline converts all the video clips into their text representations, generates a script and assembles selected video clips accordingly. 
    }
    \label{fig:video_assembly_pipeline}
\vspace{-0.4cm}
\end{figure}

However, deploying VLM pipelines on mobile devices presents fundamental challenges due to both memory and latency constraints. 
In standard pipelines, subtasks such as video encoding, caption generation, and script generation are handled by separate models~\cite{yang2023shotretrievelandassembly, xu2023retrievalbasedvideolanguagemodel}, each requiring its own memory footprint and processing time. 
For instance, even after applying 8-bit quantization, \textsf{mPLUG2}~\cite{xu2023mplug2}, a model commonly used for video-to-caption generation, requires approximately 5.6GB of RAM. Adding a script-generation model such as \textsf{Llama 3.2} can consume an additional 1.4GB. Running the full pipeline would therefore require around 7GB of memory, which can strain mobile devices that typically have only 6 to 8GB of RAM.
As a result, models used in the pipeline are loaded sequentially at each stage, as shown in Figure~\ref{fig:video_assembly_pipeline}. 
This back-to-back loading prohibits pipelined execution across tasks and inflates end-to-end latency. 
Empirically, we observe up to a 17\% increase in latency when reloading quantized versions of these models between subtasks.
This inefficiency not only prolongs wait time but also limits scalability across multiple user prompts, where each prompt restarts the model-loading cycle.

\begin{figure*}[!h]
     \centering
     \begin{subfigure}[b]{0.49\textwidth}
         \centering
         \includegraphics[width=\linewidth]{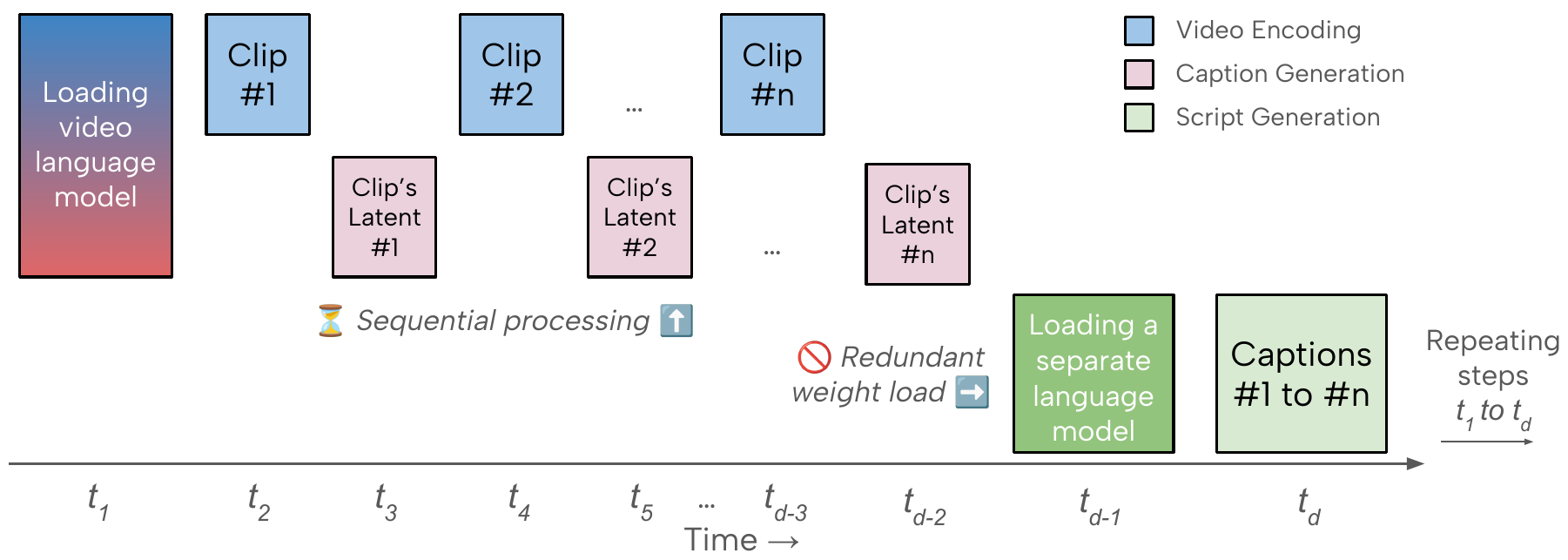}
        \caption{Sequential execution in de facto pipelines. 
        Subtasks are executed one after another, with separate models loaded at each stage (e.g., at $t_1$ and $t_{d-1}$), resulting in higher latency and memory overhead. \\
        }
         \label{fig:naive-pipeline}
     \end{subfigure}
     \hfill
     \begin{subfigure}[b]{0.49\textwidth}
         \centering
         \includegraphics[width=\linewidth]{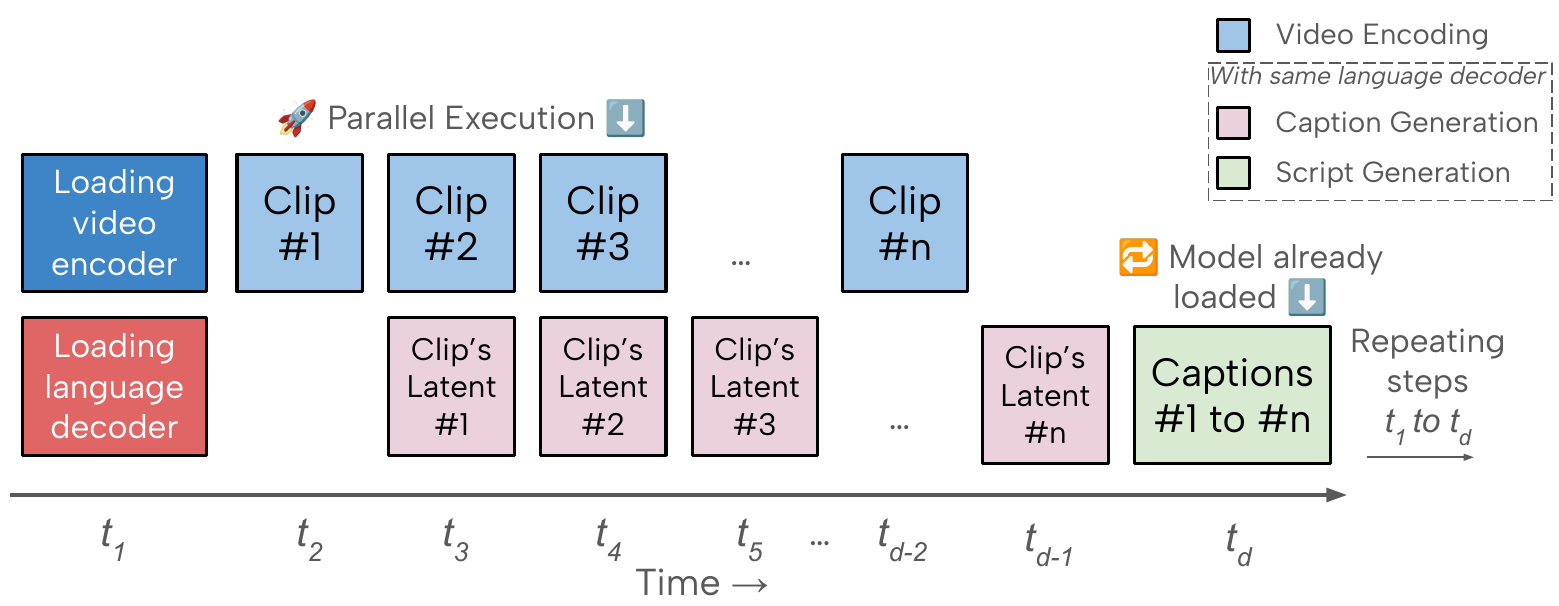}
        \caption{
        Modular, reuse-centric, and parallel execution in \projectname. 
        It reduces pipeline latency by modularizing the video encoder and language decoder; and parallelizing their execution across different inputs (e.g., starting at $t_3$). 
        }
         \label{fig:atom-pipeline}
     \end{subfigure}
        \caption{Sequential (\textit{left}) vs. \projectname's modular-parallel (\textit{right}) pipeline execution.}
        \label{fig:pipeline-comparison}
        \vspace{-0.3cm}
\end{figure*}

Prior works have predominantly optimized individual model performance: accelerating token generation or reducing memory consumption per model~\cite{chu2023mobilevlm, li2023spotlight, mllm2024xu}. 
While effective in improving pipeline performance, such efforts overlook \textit{pipeline-level inefficiencies} inherent to chaining multiple subtasks. 
In a conventional video assembly pipeline (Figure~\ref{fig:naive-pipeline}), all video clips are first passed through the video encoder and caption generator before a separate script generator is invoked. 
This serialized execution means (1) the shared VLM cannot process the next batch of video clips while script generation is still running, and 
(2) the use of a standalone script-generation model imposes a memory tradeoff: 
either keep both models in memory (which may exceed device limits) or pay the overhead of loading them sequentially. 
These limitations amplify latency and hinder responsiveness on edge devices. 
Overcoming them requires a modular, reuse-centric execution model that enables parallelism across inputs and subtasks.

\textbf{Our Approach.}
We propose \projectname, a reuse-centric inference system that accelerates video-language pipelines by addressing the pipeline-level inefficiencies. 
Unlike the sequential baseline, which uses separate models for each subtask in the pipeline, 
our approach leverages a single modularized model architecture that supports all subtasks, enabling end-to-end reuse without compromising model quality.
Specifically, 
\projectname decomposes the VLM into persistent and reusable modules: 
a video encoder and a generalizable language decoder, both of which remain resident in memory throughout the application’s lifetime. 
By reusing these modules across subtasks and avoiding repeated model loading, \projectname eliminates data movement overhead and enables parallel execution without increasing memory usage.
This reuse-centric design is compatible with existing model compression techniques, offering additional opportunities to reduce memory consumption.

Figure~\ref{fig:pipeline-comparison} contrasts this reuse-centric design with the de facto multi-model baseline. 
In the status quo (Figure~\ref{fig:naive-pipeline}), each video clip triggers a strictly serial workflow: 
the device first loads a monolithic encoder-decoder pair to generate captions sequentially for each video clip and then bring in a separate language model for script generation. 
Our approach (Figure~\ref{fig:atom-pipeline}) decomposes a VLM into separately callable encoder and decoder modules that operate concurrently: 
while the decoder produces captions for clip \#1, the encoder can already process clip \#2. 
Then the same decoder is used for processing all the captions for generating a script for video assembly. 
This modular parallelism reduces cumulative model-loading time and boosts hardware utilization without increasing the memory footprint, delivering substantial end-to-end speed-ups on mobile devices.

\textbf{Empirical Results.} 
We evaluate \projectname on two representative video-language tasks (video retrieval and video assembly) 
on commercial mobile devices including the Google Pixel 5a, Pixel 8a, and Samsung Galaxy S23. 
Compared to conventional multi-model pipelines, \projectname 
(a) reduces model loading latency by up to 50\%; 
(b) achieves up to 33\% lower end-to-end latency through reuse and parallelism;
(c) maintains competitive output quality with only marginal differences in performance metrics (e.g., 1.3–2.3 drop in Recall@$k$ for retrieval, 0.6–1.5 drop in CIDEr for captioning); and
(d) keeps peak memory virtually unchanged (only 0.76\% $\approx$ 40MB higher than the baseline) while using a stronger, reusable text decoder that still fits within a 6GB smartphone RAM budget.
While our evaluation focuses on video retrieval and video assembly, the reuse-centric pipeline design introduced by \projectname is broadly applicable to a wide range of video-language applications, including summarization, question answering, and temporal event localization.

This work makes the following contributions:
\begin{itemize}[noitemsep, nolistsep]
\item We propose a reuse-centric pipeline design that highlights a new system-level direction for efficient video-language inference. The design accelerates video-language inference by sharing a single model’s internal modules across subtasks. 
This eliminates the overhead of managing multiple models during execution. 
\item We develop \projectname, a mobile inference system that enables the execution of full video-language pipelines on-device without relying on server infrastructure.
\item We evaluate \projectname on real-world smartphones and demonstrate significant latency reductions and system-level efficiency improvements while preserving high output quality across tasks.
\end{itemize}

\section{Related Works}
\label{sec:related_works}
We organize related work into two categories: 
(1) on-device inference for vision-language models (VLMs) and their deployment strategies, and 
(2) pipelines for video-centric tasks such as assembly and retrieval. 
\projectname bridges these areas through a reuse-centric design that improves end-to-end latency for video-language pipelines on mobile devices.

\textbf{On-device Vision-Language Models.}
Recent work has explored adapting VLMs for mobile hardware. 
\textsc{MobileVLM}~\cite{chu2023mobilevlm} and \textsc{Spotlight}~\cite{li2023spotlight} demonstrate that billion-parameter VLMs can run efficiently on smartphones by pairing lightweight visual encoders with language decoders. 
However, these efforts primarily target image-based tasks and assume single-stage inference. 
They do not address the challenge of executing multi-stage pipelines, where repeated model loading and lack of reuse can introduce substantial latency overhead.
\projectname complements these efforts by focusing on architectural reuse within the pipeline to accelerate full-stack video-language inference on-device, rather than optimizing individual models in isolation.

\textbf{Video-Centric Pipelines.}
Video-language tasks such as assembly and retrieval typically rely on modular pipelines composed of separate models for video encoding, captioning, and reasoning~\cite{yang2023shotretrievelandassembly, xu2023retrievalbasedvideolanguagemodel}. 
While effective, this approach incurs high runtime overhead, particularly on mobile devices, due to the need to repeatedly load and execute distinct models for each subtask into limited memory.
For video assembly, methods like \textsc{RATV}~\cite{yang2023shotretrievelandassembly} and \textsc{TV-MGI}~\cite{yin2024textvideomultigrainedintegrationvideo} align video segments to human-written scripts using powerful server-hosted models. 
These approaches assume expert-written inputs and high-resource environments. 
\projectname, by contrast, automates both script generation and clip selection directly on-device using a unified model, making the process more accessible to end users.
In video retrieval, prior work such as \textsc{CLIP4Clip}~\cite{luo2021clip4clip} and \textsc{Frozen}~\cite{bain21frozen} relies on embedding-based alignment between video and text. 
More recent methods~\cite{xu2023retrievalbasedvideolanguagemodel, jeong2025videoragretrievalaugmentedgenerationvideo} integrate LLM-based reasoning to enhance retrieval quality, but depend on large proprietary models and cloud infrastructure. 
These methods are not deployable in resource-constrained or privacy-sensitive settings. 
\projectname shows that retrieval, assembly, and captioning can all be supported within a single reusable model on mobile devices, eliminating the need for cloud-based processing or manual supervision.

\section{Design of \projectname}
\label{sec:design}

\projectname accelerates video-language pipeline inferences by eliminating pipeline-level redundancies through three key design principles:
\begin{itemize}[]
    \item \textit{Modularization:} A VLM is split into a video encoder and a stronger, reusable language decoder, each independently callable for reuse and concurrent execution.
    \item \textit{Reuse:} Modules are persistently kept in memory and shared across subtasks, avoiding repeated loading.
    \item \textit{Parallelism:} Decoupled modules enable thread-safe parallel execution, improving throughput and device utilization.
\end{itemize}

\subsection{Modular Execution of VLM Pipelines}
\label{subsec:modular-execution}

In traditional monolithic VLM pipelines~\cite{yang2023shotretrievelandassembly}, tasks such as video captioning and script generation (shown in Figure~\ref{fig:video_assembly_pipeline})
are implemented using separate models or tightly coupled end-to-end models. 
This results in high latency due to serialized execution and repeated model reloading. 
Moreover, monolithic VLMs prevent concurrent execution of different subtasks.

\projectname addresses these inefficiencies through modular decomposition. 
The video encoder and text decoder components of the VLM are exposed as independently callable modules. 
This allows concurrent execution, for example, encoding a new video clip while simultaneously decoding captions from prior clips. 
A further benefit of modularization is that the \emph{same} high-capacity language decoder can be reused across all text-centric stages.  
Replacing multiple specialized, lighter decoders with a single shared decoder
(i) eliminates repeated model loading,
(ii) promotes generalization by applying a stronger linguistic prior across all text-centric subtasks (e.g., captioning, retrieval indexing, script generation), and
(iii) crucially, enables the entire pipeline to run within tight mobile memory budgets.
In contrast, pipelines that tries to load together separate models for captioning and script generation often exceed the 6GB RAM limit on common smartphones.
Our reuse of a single high-capacity decoder avoids this constraint while still improving quality.
Empirical comparisons are reported in Section~\ref{subsec:performance}.

\subsection{Persistent Module Reuse}
\label{subsec:reuse-strategy}

Even with modularization, sequential implementations that rely on task-specific models may reload different modules at each stage of a multi-stage pipeline, wasting compute and memory bandwidth.
\projectname avoids this overhead through a \textit{persistent reuse strategy}, keeping key modules; such as the video encoder and a stronger, unified language decoder, loaded in memory across tasks.
For example, the same decoder used for caption generation is also reused for script composition, eliminating the need to load a separate language model for the latter stage.

This reuse strategy reduces latency and avoids memory thrashing under constrained resources. 
As shown in Figure~\ref{fig:naive-pipeline}, non-reuse pipelines incur significant delays from repeated model loading and unloading, especially when using large language decoders. 
In contrast, \projectname (Figure~\ref{fig:atom-pipeline}) retains modules in memory throughout execution, streamlining transitions across subtasks and ensuring smoother operation on mobile hardware.

Although the memory footprint of model weights and activations is comparable between implementations, reuse-centric design lowers auxiliary memory usage (e.g., temporary allocations during weight loading). 
The detailed memory breakdown is provided in Appendices~\ref{adx:memory-footprint} and~\ref{subsec:layerwise-memory-breakdown}.

\subsection{Parallelism through Modular Design}
\label{subsec:parallelism}

A key advantage of modularizing the VLM is enabling \textit{parallel execution} of subtasks, something infeasible with monolithic or tightly coupled models.
Traditional video-language pipelines enforce serialized execution due to thread-unsafe or memory-intensive models: 
caption generation must wait for video encoding to finish, and script generation must wait for captioning to complete.
This leads to under-utilization of compute resources and increased end-to-end latency.

In contrast, \projectname's decoupled encoder and decoder are exposed as independently callable, thread-safe modules that persist in memory throughout execution.
This persistent modular design allows pipeline stages to overlap: 
for example, while the decoder generates captions for clip~\#1, the encoder can begin processing clip~\#2.
Figure~\ref{fig:atom-pipeline} illustrates how overlapping compute windows boosts throughput and improves hardware utilization on edge devices.

While this parallel execution increases peak activation memory, we show in Appendix~\ref{adx:memory-footprint} that \projectname offsets this by eliminating temporary memory overhead from repeated model loading, resulting in only 40MB higher overall usage.

\paragraph{Applicability to Other Tasks.}
While our analysis focuses on video retrieval and assembly for concreteness, the modular and reuse-centric design in \projectname generalizes to other video-language tasks, including summarization, and question answering. 
We demonstrate this with additional task implementation outlines described in Appendix~\ref{adx:broader_applications}.

\section{Implementation of \projectname} 
\label{sec:implementation-of-atom}
We implement \projectname on top of the vision-language foundation model \textsf{mPLUG2}~\cite{xu2023mplug2}. We modularize its components to support persistent module reuse and parallel execution, enabling efficient on-device inference. 

We choose \textsf{mPLUG2} for three reasons: 
(a) it has demonstrated strong performance in video-language tasks like captioning and retrieval~\cite{wu2024next, chen2024panda, wang2024internvideo2}, 
(b) it is open-source and modular, allowing targeted architectural changes, and 
(c) it is compatible with quantization and serialization frameworks such as \textsf{TorchAO}~\cite{torchao} for mobile deployment.
The original \textsf{mPLUG2} consists of a \textsf{BERT}-based text encoder and decoder (0.3B parameters each), a \textsf{CLIP-ViT/L-14} visual encoder (0.4B), and a fusion module (0.5B), totaling 1.5B parameters. 
However, it is not optimized for task-level reuse: 
the decoder requires separate fine-tuning for different language tasks (e.g., captioning vs. script generation), increasing training and runtime overhead.

To enable reuse across subtasks, we make two key modifications. 
First, we discard the text encoder and fusion module, which are not required for our tasks. Second, we replace the \textsf{BERT}-based decoder with a 1B parameter \textsf{Llama 3.2}-based autoregressive decoder~\cite{grattafiori2024llama3herdmodels}. 
This decoder generalizes across captioning and reasoning tasks, allowing the same module to serve both roles without retraining.

The resulting model, which we refer to as \textsf{mPLUG2+}, consists of a CLIP-based video encoder and a \textsf{Llama}-based text decoder, totaling 1.4B parameters. 
We fine-tune the model on video captioning datasets to align the visual and text components; script generation requires no additional training. 
Compared to the original \textsf{mPLUG2}, this update improves captioning (by 2.8–3.7 CIDEr) and retrieval (by 3.8–4.9 Recall@1) performance, as shown in Section~\ref{subsec:performance}.
To support reuse-centric execution, each module is serialized independently and kept persistently loaded during pipeline execution. 
This design enables both shared reuse and thread-safe parallelism across subtasks, enabling reliable execution on mobile devices with 6GB+ RAM.

\paragraph{Adaptability to Other VLMs.} While our implementation uses \textsf{mPLUG2+} as the base model, \projectname's reuse-centric design is broadly compatible with many modern VLMs that follow a modular architecture, such as BLIP-2, Flamingo, and VideoCoCa. 
These models decouple video encoders and language decoders, making them suitable for persistent reuse. 
Although earlier tightly coupled models with joint cross-modal attention layers may resist such decomposition, deployment on mobile devices generally necessitates modularization for latency and memory efficiency, an assumption increasingly reflected in newer model architectures.

\paragraph{Quantization.}
To support deployment on memory-constrained mobile hardware, we apply model quantization to reduce the memory footprint. 
We apply dynamic quantization using \textsf{TorchAO} to enable efficient deployment of our video-language models on mobile devices. 
Focusing on Linear and Embedding layers (responsible for over 91\% of memory use), we achieve int8 compatibility and serialization for ARM CPUs while avoiding unreliable Conv3D quantization.
Details of quantization, operator compatibility, serialization, and deployment constraints are included in Appendix~\ref{adx:quantization} and~\ref{adx:serialization}. 

\section{Results and Discussion}
\label{sec:results-and-discussion}
We evaluate \projectname on two representative video-language pipelines (video assembly and video retrieval) demonstrating the following key findings:
(a) The modified \textsf{mPLUG2+} architecture supports task reuse without performance degradation; and
(b) \projectname's reuse-centric design substantially reduces end-to-end latency by avoiding repeated weight loading and enabling parallelism.

We begin by detailing our experimental setup in Section~\ref{subsec:experimental-setup}, followed by results and analysis in Sections~\ref{subsec:performance} and \ref{subsec:latency}. 
Additional qualitative examples and results are provided in Appendix~\ref{adx:qualitative-analysis} and~\ref{adx:additional-results}.

\subsection{Experimental Setup}
\label{subsec:experimental-setup}

\paragraph{Tasks and Models.} 

We choose video retrieval and video assembly to represent two ends of the video-language task spectrum: 
retrieval-based reasoning and generative storytelling. 
Despite their differences, both benefit from caption generation as a shared intermediate step.
This enables us to reuse the same language decoder across tasks, reducing latency and memory spent in the model loading phase. 
While server-side systems may afford task-specific models, our design favors reuse to meet mobile constraints. 
Appendix~\ref{adx:broader_applications} discusses how this approach extends to other tasks like summarization.

Both pipelines rely on three core modules: a visual encoder for video encoding, a language decoder for caption generation (conditioned on video features), and a sentence transformer for matching user prompts to generated captions. The language decoder is also reused for script generation, enabling efficiency in \projectname's reuse-centric design.

As described in Section~\ref{sec:implementation-of-atom}, we use \textsf{CLIP-ViT/L-14} (0.4B parameters) as the visual encoder and \textsf{Llama 3.2} (1B parameters) as the language decoder. The model is fine-tuned for video captioning using four NVIDIA A100 GPUs and the datasets described below. For semantic matching, we use \textsf{all-MiniLM-L6-v2}~\cite{sentencetransformerminilm}, a 22.7M parameter sentence transformer.

\paragraph{Datasets.}
We use three datasets to fine-tune and evaluate the modified \textsf{mPLUG2+} for video captioning, video retrieval, and qualitative video assembly (Appendix~\ref{adx:qualitative-analysis}):
(a)~\textsf{MSR-VTT}~\cite{xu2016msr-vtt},
(b)~\textsf{MSVD}~\cite{chen2011msvd}, and
(c)~\textsf{DiDeMo}~\cite{hendricks2018didemo}.
The model is trained on the combined training splits of all three datasets and evaluated on their respective test sets. The first two were also used in training the original \textsf{mPLUG2}.

\textsf{MSR-VTT} contains 10K YouTube videos with 200K captions, split into 6.5K training, 0.5K validation, and 3K test samples.
\textsf{MSVD} has 1,970 YouTube clips with captions, divided into 1.2K/100/670 training/val/test samples.
\textsf{DiDeMo} includes 10K Flickr videos, split into 8K/1K/1K, each paired with four text descriptions.

\paragraph{Counterparts for Comparison.}
Prior works on video assembly~\cite{yang2023shotretrievelandassembly, yin2024textvideomultigrainedintegrationvideo} rely on human-written scripts to retrieve relevant clips. In contrast, \projectname generates scripts from video clips based on user prompts. These prior systems are also closed-source, preventing reproducible comparisons.
For video retrieval, existing methods focus on 
(a) architectural advances~\cite{xu2023retrievalbasedvideolanguagemodel, xu2023mplug2} or 
(b) improving retrieval relevance~\cite{tevissen2024ragforlargevideo}, but those designed for mobile settings~\cite{zhang2021personalizedvideoretrievalmobiledevices, tahboub2015contentbasedvideoretrieval} do not leverage large language models for reasoning. 

Due to these limitations, we evaluate \projectname using system-level baselines focused on performance and efficiency:
(a)~a 96-core x86\_64 CPU server, both with and without \projectname's reuse-centric design;
(b)the same server equipped with an NVIDIA A40 GPU, evaluated in both quantized and full-precision modes, again with and without reuse; and
(c)~\projectname on mobile devices without reuse, where separate models are loaded and unloaded for each subtask.
The third baseline reflects typical multi-stage pipelines without coordination. 
Comparing against it isolates the benefits of \projectname's modular reuse in latency and memory efficiency.

\paragraph{Metrics.}
For evaluating \textbf{video captioning subtask} performance, we employ four metrics: 
(a)~BLEU@4, 
(b)~ROUGE-L, 
(c)~METEOR, and 
(d)~CIDEr. 
BLEU@4 calculates the geometric mean of $n$-gram precision up to 4-grams between generated and reference captions, applying a brevity penalty. 
ROUGE-L assesses content overlap and word order by finding the longest common subsequence. 
METEOR computes a weighted harmonic mean of precision and recall, incorporating various linguistic features. 
CIDEr measures similarity using TF-IDF weighted cosine similarity of $n$-grams, emphasizing distinctive phrases. 
These metrics collectively provide a comprehensive assessment of caption quality, considering factors such as precision, recall, semantic similarity, and distinctiveness.

For evaluating \textbf{video retrieval task} performance, we use 
(a)~Recall@1, 
(b)~Recall@5, and
(c)~Recall@10.
Recall@$k$ measures the proportion of relevant videos successfully retrieved within the top $k$ results of the video retrieval pipeline.

Meanwhile, the evaluation of \textbf{video assembly} has more human factors involved and there are no established metrics for the task.
Hence we have performed qualitative analysis to measure efficacy of \projectname, as shown in Appendix~\ref{adx:qualitative-analysis}. 
For the task of \textbf{semantic matching} through sentence transformer, the metric of Pearson correlation measures the linear relationship between two sentence embeddings, quantifying semantic similarity based on how closely their values co-vary.

\textbf{Latency} is measured in seconds. 
We have measured the latency of loading model(s), running each component or a subtask (video encoding, caption generation) of a pipeline, as well as the latency of running the end-to-end pipeline. 
\textbf{Memory consumption} is measured in Gigabytes (GBs). 

\paragraph{Hardware.}
We evaluated \projectname on 3 physical mobile devices with diverse computing capabilities:
(a)~Google Pixel 5a (6GB RAM),
(b)~Google Pixel 8a (8GB RAM), and 
(c)~Samsung Galaxy S23 (8GB RAM).
We have used \textsf{Pytorch Mobile} as the inference engine.
Built for \textsf{PyTorch Mobile}, we also utilized \textsf{TorchAO} for optimized execution through operator fusion, quantization, and lightweight runtime adaptations.
The experiments focused on executing the end-to-end task pipelines on mobile CPUs, given the limited compatibility of \textsf{TorchAO} (this limitation has been observed in other works as well~\cite{mllm2024xu, xu2023llmcadfastscalableondevice}) with alternative processing units such as GPUs and NPUs. 
However, \projectname remains adaptable to other processing units, making it orthogonal to the choice of hardware accelerators.

\paragraph{Hyperparameters.}
We follow the hyperparameters used for training \textsf{mPLUG2}~\cite{xu2023mplug2}.
The training lasts for 10 epochs, with the learning rate of 2e-5 for the task of video description.
The batch size is set to 128.
The max caption generation length is set to 23 tokens, which is needed for fixing the  decoding loop iterations of JIT tracing.
Beam size of 4 is used for the caption generation.
The videos are preprocessed to have sampling rate of 6 frames per second, and each frame's resolution is decreased to 224$\times$224.
The hyperparameter selection is discussed in Appendix~\ref{adx:hyperparameter-selection}.

\subsection{Performance}
\label{subsec:performance}
In this section, we demonstrate the advantages of integrating a stronger language decoder in on-device video-language pipelines for retrieval and captioning tasks.
Our enhanced \textsf{mPLUG2+}, powered by a \textsf{LLaMa} text decoder, delivers substantial gains in prediction performance compared to the original \textsf{mPLUG2} with a \textsf{BERT} decoder, achieving 2.8--3.7 and 6.8--7.2 CIDEr point improvements without and with quantization, respectively.
Moreover, quantization has a minimal impact on performance, causing only a 0.6--1.3 CIDEr reduction, while enabling efficient deployment on mobile and edge devices.
These advancements allow us to build the first fully operable on-device video assembly pipeline, making real-time video understanding feasible.

We first discuss the performance of video captioning and video retrieval tasks, as detailed in Tables~\ref{tab:captioning-performance} and \ref{tab:retrieval-performance} respectively. 
Due to the lack of quantitative metrics for the video assembly task, we have conducted a qualitative analysis on the assembled videos with 10 human participants, those results are included in Appendix~\ref{adx:qualitative-analysis}. 
\begin{table}[!t]
\centering
\footnotesize
\addtolength{\tabcolsep}{-0.2em}
\begin{tabular}{l|cccc|cccc}
\toprule
\multirow{2}{*}{\makecell{Model\\Variant}} & \multicolumn{4}{c|}{MSR-VTT} & \multicolumn{4}{c}{MSVD} \\
\cmidrule{2-9}
 & B & R & M & C & B & R & M & C \\
 \midrule
 & \multicolumn{8}{c}{Without Quantization (32-bit weights)} \\
 \midrule
\begin{tabular}[c]{@{}l@{}}\textsf{mPLUG2} \end{tabular} & 56.9 & 34.0 & 68.2 & 79.4 & 73.4 & 46.8 & 84.7 & 163.6 \\
\begin{tabular}[c]{@{}l@{}}\textsf{mPLUG2+} \end{tabular} & 62.4 & 36.8 & 70.8 & 83.1 & 75.3 & 49.1 & 86.8 & 166.4 \\
\midrule
 & \multicolumn{8}{c}{With Quantization (8-bit weights)} \\
 \midrule
\begin{tabular}[c]{@{}l@{}}\textsf{mPLUG2} \end{tabular} & 52.4 & 30.8 & 65.4 & 75.3 & 70.8 & 43.2 & 80.7 & 158.1 \\
\begin{tabular}[c]{@{}l@{}}\textsf{mPLUG2+}\end{tabular} & 60.7 & 35.9 & 68.6 & 82.5 & 74.2 & 46.3 & 84.5 & 164.9 \\
\bottomrule
\end{tabular}
\caption{\textbf{Performance} comparison of the original \textsf{mPLUG2} (with \textsf{BERT} text decoder) and \textsf{mPLUG2+} (with \textsf{LLaMa} text decoder) for both quantized and full-precision versions for the task of \textbf{video captioning}. 
All 4 metrics B (BLEU@4), R (ROUGE-L), M (METEOR), and C (CIDEr) are ``higher the better''.
\textsf{mPLUG2+} of our method \projectname has a lower performance drop even after quantization, compared to the original \textsf{mPLUG} with a weaker text decoder.}
\label{tab:captioning-performance}
\vspace{-0.2cm}
\end{table}

Compared to the original \textsf{mPLUG2}, our modified \textsf{mPLUG+} improves BLEU@4 by 1.9--5.5 points, indicating better alignment between generated captions and reference captions. 
It also achieves ROUGE-L gains of 2.3--2.8 points, reflecting improved phrase-matching accuracy, and METEOR gains of 2.1--2.6 points, which capture better word alignment. 
Furthermore, CIDEr, which measures caption informativeness, improves by 2.8--3.7 points. 
Even after quantization, \textsf{mPLUG2+} consistently outperforms the original model by 6.8--7.2 CIDEr score.
Due to quantization, we see a minor degradation of 0.6--1.5 CIDEr score compared to full-precision \textsf{mPLUG+}, while maintaining a negligible qualitative difference, as detailed in Appendix~\ref{adx:qualitative-analysis}.

Similarly, our retrieval results in Table~\ref{tab:retrieval-performance} show consistent performance improvements across MSR-VTT, and MSVD datasets.
Extended results on DiDeMo dataset are available in Appendix~\ref{adx:extended-results-didemo}.
The full-precision \textsf{mPLUG2+} model achieves Recall@1 gains of 3.8--4.9 points over \textsf{mPLUG}, demonstrating its stronger ability to retrieve the most relevant video. 
We observe similar trends for Recall@5 and Recall@10, reinforcing the robustness of our model across different ranking positions. 
Even with quantization, \textsf{mPLUG2+} maintains a retrieval advantage over the original \textsf{mPLUG2}, with a Recall@1 drop of only 1.3--2.3 points, yet still outperforming the quantized \textsf{mPLUG} model by 4.7--7.1 points.
Additionally, we show (in Appendix~\ref{subsec:sentence-transformer}) that the \textsf{all-MiniLM-L6-v2} sentence transformer for caption-to-prompt matching in both retrieval and assembly pipelines has negligible impact on the end-to-end quality of our system, even after quantization. This ensures efficient and accurate matching in resource-constrained settings.
\begin{table}[!t]
\centering
\footnotesize
\addtolength{\tabcolsep}{-0.1em}
\begin{tabular}{l|ccc|ccc}
\toprule
\multirow{2}{*}{\makecell{Model\\Variant}} & \multicolumn{3}{c|}{MSR-VTT} & \multicolumn{3}{c}{MSVD} \\ \cmidrule{2-7}
 & R@1 & R@5 & R@10 & R@1 & R@5 & R@10 \\ \midrule
\multicolumn{7}{c}{Without Quantization (32-bit floating point weights)} \\ \midrule
\textsf{mPLUG2} & 52.6 & 75.8 & 83.1 & 65.3 & 82.0 & 90.6 \\
\textsf{mPLUG2+} & 57.2 & 80.7 & 85.1 & 69.1 & 84.5 & 93.8 \\ \midrule
\multicolumn{7}{c}{With Quantization (8-bit integer weights)} \\ \midrule
\textsf{mPLUG2} & 48.7 & 72.6 & 80.5 & 62.1 & 77.3 & 87.9 \\
\textsf{mPLUG2+} & 55.8 & 78.4 & 82.9 & 66.8 & 82.7 & 90.3 \\ \bottomrule
\end{tabular}
\caption{\textbf{Performance} comparison of the original \textsf{mPLUG2} (with \textsf{BERT} text decoder) and our modified \textsf{mPLUG2+} (with \textsf{Llama} text decoder) for both quantized and full-precision versions for the task of \textbf{video retrieval}. 
All 3 metrics R@1 (Recall@1), R@5 (Recall@5), and R@10 (Recall@10) are ``higher the better''.
\textsf{mPLUG2+} of \projectname has a lower performance drop even after quantization, compared to the original \textsf{mPLUG} with a weaker text decoder.
}
\vspace{-0.2cm}
\label{tab:retrieval-performance}
\end{table}

\subsection{Latency}
\label{subsec:latency}
This section presents how the reuse-centric design for efficient pipeline execution of \projectname leads to latency reduction of 27--33\% for video assembly and video retrieval tasks.

Table~\ref{tab:captioning-retrieval-latency} shows that \projectname achieves substantial latency reductions across resource-constrained devices. 
On the Google Pixel 5a, our reuse-centric design lowers the end-to-end video retrieval latency by 33.06\%, and video assembly latency by 30.78\%. 
Similar trends are observed on the Google Pixel 8a and Samsung Galaxy S23, with latency reductions of approximately 30.43\% and 29.14\% for retrieval, and 27.69\% and 27.54\% for assembly, respectively. 
These reductions are driven by the parallel execution and amortized reuse of computational modules, significantly reducing redundant processing.

\begin{table*}[!t]
\fontsize{7.5pt}{9pt}\selectfont
\addtolength{\tabcolsep}{-0.3em}
\begin{tabular}{l|lc|cccc|cc}
\toprule
 & \multicolumn{1}{l|}{} & \multicolumn{1}{l|}{} & \multicolumn{4}{c|}{\makecell{Component-wise*}} & \multicolumn{2}{c}{End-to-end} \\ \midrule
 & \multicolumn{1}{l|}{Platform Variant} & \multicolumn{1}{c|}{\makecell{Model\\Loading}} 
 & \makecell{Video\\Encoding\\(CLIP-ViT)} 
 & \makecell{Caption\\Generation\\(Llama)} 
 & \makecell{Indexing\\(MiniLM)} 
 & \makecell{Script\\Generation\\(ST + Decoder)} 
 & \makecell{Video\\Retrieval} 
 & \makecell{Video\\Assembly} \\ 
\midrule 
\multirow{18}{*}{\begin{tabular}[c]{@{}l@{}}High-end\\ Hardware\end{tabular}} & \multicolumn{8}{c}{Without Quantization} \\ \cmidrule{2-9} 
 & \multicolumn{1}{l|}{x86\_64 CPU (w/o reuse)} & \multicolumn{1}{c|}{6.19} & 1.68 & 1.15 & 10.35 & \multicolumn{1}{c|}{11.09} & 36.05 & 40.67 \\
 & \multicolumn{1}{l|}{\begin{tabular}[c]{@{}l@{}}x86\_64 CPU + GPU (w/o reuse)\end{tabular}} & \multicolumn{1}{c|}{8.66} & 0.15 & 0.08 & 6.48 & \multicolumn{1}{c|}{6.84} & 12.12 & 14.95 \\ \cmidrule{2-9} 
 & \multicolumn{1}{l|}{x86\_64 CPU (\textbf{\projectname})} & \multicolumn{1}{c|}{3.44 {\scriptsize ($\downarrow$ 44.43\%)}} & 1.68 & 1.15 & 10.35 & \multicolumn{1}{c|}{11.09} & 26.58 ({\scriptsize $\downarrow$ 26.27\%}) & 30.28 ({\scriptsize $\downarrow$ 25.55\%}) \\
 & \multicolumn{1}{l|}{\begin{tabular}[c]{@{}l@{}}x86\_64 CPU + GPU (\textbf{\projectname})\end{tabular}} & \multicolumn{1}{c|}{4.38 {\scriptsize ($\downarrow$ 49.42\%)}}  & 0.15 & 0.08 & 6.48 & \multicolumn{1}{c|}{6.84} & 8.65 ({\scriptsize $\downarrow$ 28.63\%}) & 10.46 ({\scriptsize $\downarrow$ 30.03\%}) \\ \cmidrule{2-9} 
 & \multicolumn{8}{c}{With Quantization} \\ \cmidrule{2-9} 
 & \multicolumn{1}{l|}{x86\_64 CPU (w/o reuse)} & \multicolumn{1}{c|}{7.32} & 2.95 & 2.14 & 13.66 & \multicolumn{1}{c|}{15.40} & 44.84 & 47.59 \\
 & \multicolumn{1}{l|}{\begin{tabular}[c]{@{}l@{}}x86\_64 CPU + GPU (w/o reuse)\end{tabular}} & \multicolumn{1}{c|}{8.89} & 0.46 & 0.32 & 10.78 & \multicolumn{1}{c|}{13.02} & 19.20 & 22.18 \\ \cmidrule{2-9}
 & \multicolumn{1}{l|}{x86\_64 CPU (\textbf{\projectname})} & \multicolumn{1}{c|}{3.64 ({\scriptsize $\downarrow$ 50.27\%})} & 2.95 & 2.14 & 13.66 & \multicolumn{1}{c|}{15.40} & 32.12 ({\scriptsize $\downarrow$ 28.37\%}) & 34.69 ({\scriptsize $\downarrow$ 27.11\%}) \\
 & \multicolumn{1}{l|}{\begin{tabular}[c]{@{}l@{}}x86\_64 CPU + GPU (\textbf{\projectname})\end{tabular}} & \multicolumn{1}{c|}{4.56 ({\scriptsize $\downarrow$ 48.71\%})} & 0.46 & 0.32 & 10.78 & \multicolumn{1}{c|}{13.02} & 14.02 ({\scriptsize $\downarrow$ 26.98\%}) & 16.38 ({\scriptsize $\downarrow$ 26.15\%}) \\ \midrule
\multirow{6}{*}{\begin{tabular}[c]{@{}l@{}}Resource-\\ constrained\\ Hardware\end{tabular}} & \multicolumn{1}{l|}{Google Pixel 5a (w/o reuse)} & \multicolumn{1}{c|}{32.94} & 8.68 & 5.57 & 24.39 & \multicolumn{1}{c|}{26.61} & 162.33 & 167.52 \\
 & \multicolumn{1}{l|}{Google Pixel 8a (w/o reuse)} & \multicolumn{1}{c}{29.63} & 6.34 & 4.22 & 21.20 & \multicolumn{1}{c|}{22.41} & 147.04 & 153.59 \\
 & \multicolumn{1}{l|}{Samsung Galaxy S23 (w/o reuse)} & \multicolumn{1}{c|}{28.43} & 6.83 & 3.54 & 21.03 & \multicolumn{1}{c|}{22.68} & 146.47 & 152.35 \\ \cmidrule{2-9} 
 & \multicolumn{1}{l|}{Google Pixel 5a (\textbf{\projectname})} & \multicolumn{1}{c|}{17.23 (\scriptsize $\downarrow$ 47.69\%)} & 8.68 & 5.57 & 24.39 & \multicolumn{1}{c|}{26.61} & 108.66 (\scriptsize $\downarrow$ 33.06\%) & 115.95 (\scriptsize $\downarrow$ 30.78\%) \\
 & \multicolumn{1}{l|}{Google Pixel 8a (\textbf{\projectname})} & \multicolumn{1}{c|}{15.87 (\scriptsize $\downarrow$ 46.44\%)} & 6.34  & 4.22  & 21.20 & \multicolumn{1}{c|}{22.41} & 102.30 (\scriptsize $\downarrow$ 30.43\%) & 111.06 (\scriptsize $\downarrow$ 27.69\%) \\
 & \multicolumn{1}{l|}{Samsung Galaxy S23 (\textbf{\projectname})} & \multicolumn{1}{c|}{14.12 (\scriptsize $\downarrow$ 50.33\%)}  & 6.83  & 3.54 & 21.03 & \multicolumn{1}{c|}{22.68} & 103.78 (\scriptsize $\downarrow$ 29.14\%) & 110.39 (\scriptsize $\downarrow$ 27.54\%) \\ \bottomrule
\end{tabular}
\caption{
\textbf{Latency (in seconds, lower the better)} comparison of \projectname against a sequential (w/o reuse) implementation.
The numbers in parentheses ($\downarrow$) indicate the percentage savings of \projectname over the sequential baseline.
The table reports both \textbf{component-wise latencies} of each subtask (video encoding, caption generation, indexing, and script generation) and \textbf{end-to-end latencies} for retrieval and video assembly tasks, on high-end servers and resource-constrained mobile devices. GPU: NVIDIA A40.
*\textbf{Note}: 
Component-wise latencies are measured independently per input. 
End-to-end latency, however, reflects execution over 5 inputs, showcasing the benefits of \projectname's parallel execution and amortized model reuse. 
This design significantly reduces overall runtime by loading model modules only once and enabling concurrent processing across inputs, leading to end-to-end latencies lower than the naive sum of individual component times.
}
\vspace{-0.25cm}
\label{tab:captioning-retrieval-latency}
\end{table*}
The key factor behind these improvements is \projectname's ability to avoid repeated model loading and execution overhead. 
Traditional sequential, multi-model implementations reload models for different tasks, introducing substantial latency penalties, especially on mobile devices with constrained computational power. 
By reusing modules of a large video-language model, specifically our modified \textsf{mPLUG2+}, \projectname eliminates redundant computations and allows concurrent execution of different pipeline stages, leveraging hardware parallelism more effectively. 
This leads to noticeable latency reduction for video assembly and video retrieval, where sequential execution would otherwise introduce significant bottlenecks, as shown in ``(w/o reuse)'' counterparts of the table.

Another critical insight is the comparison between high-end server configurations and mobile hardware. 
While GPUs in high-end setups, such as the NVIDIA A40, greatly reduce latency through efficient vectorized computations, mobile devices lack such accelerations. 
In these environments, optimizing execution flow is even more crucial. 
The results confirm that \projectname's reuse-focused approach is effective in bridging this performance gap, delivering end-to-end latency reductions exceeding 30\% without requiring additional specialized hardware.

Furthermore, the impact of quantization on latency underscores the importance of \projectname's design. 
While quantization generally reduces a model's memory consumption, it introduces dynamic conversion overheads that can increase latency in  implementations of video-based pipelines without reuse. 
However, \projectname mitigates this by parallelizing the execution on video encoder and language decoder, preventing performance degradation. 
The improvements are particularly evident in resource-constrained mobile device setups, where our approach prevents excessive delays caused by sequential processing of quantized operations.

Overall, these findings validate the effectiveness of \projectname's reuse-centric pipeline execution. 
By strategically leveraging modular reuse and concurrent execution, \projectname significantly accelerates video retrieval and assembly tasks, making real-time video processing feasible even on mobile hardware. 
The results highlight the importance of optimizing computational reuse to achieve low-latency, high-efficiency AI-driven applications.

\subsection{Additional Results}
\label{subsec:additional-results}
To further support the practical viability of \projectname, Appendix~\ref{adx:additional-results} provides results covering latency-performance trade-offs, storage feasibility, and memory efficiency.

(a)~Ablation studies on a Google Pixel 5a in Appendix~\ref{subsec:ablation-studies} demonstrate the trade-offs between latency and performance under different execution configurations, including video resolution, caption length, frame rate, beam size, and batch size. These allow practitioners to adapt the system based on their specific deployment constraints.

(b)~We also analyze model storage feasibility in Appendix~\ref{subsec:storage-size}. 
The modularized video-language model, including the SentenceTransformer, occupies $\sim$1.3GB in total; making it practical for deployment on typical smartphones with 6–8GB of RAM.

(c)~Finally, Appendix~\ref{adx:memory-footprint} and~\ref{subsec:layerwise-memory-breakdown} provides a detailed memory breakdown of \projectname compared to a non-reuse variant, and a discussion on latency-memory trade-off.
Because \projectname holds the encoder and decoder in memory while they execute in parallel, activation footprint rises by 182MB, but this is offset by the elimination of temporary buffers created during repeated model loading. 
The net effect is a negligible 0.76\% ($\approx$ 40MB) increase in peak RAM, leaving the complete pipeline well within the 6GB budget of commodity smartphones.

\section{Conclusion}

In this work, we present \projectname, an efficient on-device system that enables end-to-end video-language pipelines on commodity mobile devices. 
\projectname adopts a reuse-centric design that modularizes and repurposes components of a single VLM to serve all stages of the pipeline. 
This approach not only addresses critical challenges of model loading latency, but also ensures practical deployment on devices with as little as 6GB of RAM. 
Our evaluations across commercial smartphones demonstrate that -- compared to de facto sequential, multi-model pipelines with no modularization or reuse -- \projectname achieves stronger performance (up to 3.7 CIDEr score for assembly and 4.9 Recall@1 for retrieval) on downstream tasks with minimal degradation due to quantization, while significantly improving runtime efficiency (up to 33\%). 
The qualitative evaluation echoes the quantitative gains of \projectname. 
By eliminating the need for external servers, \projectname enhances privacy, reduces infrastructure costs, and makes advanced video-language applications accessible even in bandwidth-constrained or offline environments. 

\clearpage
\appendix
\section{Use Case: Video Assembly}
\label{adx:video_assembly}

We detail how \projectname executes the video assembly task entirely on-device, showcasing its modular and reuse-centric design. This example highlights the pipeline architecture and the interaction between the video encoder and language decoder modules.

\subsection{Pipeline Overview}

Given a user prompt (e.g., “make a highlight reel”) and a set of user-provided videos, the pipeline proceeds as follows:

\begin{enumerate}
\item \textbf{Video segmentation.} Input videos are divided into overlapping clips of fixed duration ($\ell = 10s$) with a stride of $s = 5s$, enabling finer temporal granularity and reducing memory load per inference call.
\item \textbf{Video captioning.} Each clip is passed through the VLM in a two-stage process:
\begin{itemize}
    \item The video encoder extracts high-dimensional visual embeddings.
    \item The text decoder generates a concise natural language caption from the visual embedding.
\end{itemize}

\item \textbf{Embedding and ranking.} Captions and the user prompt are embedded using a Sentence Transformer. Clip relevance is computed via cosine similarity between caption and prompt embeddings.

\item \textbf{Script generation.} The top-$k$ relevant captions are composed into a coherent video script using the same VLM text decoder, now operating in text-only mode. This script determines the selected clip order and narration.

\item \textbf{Final assembly.} Selected clips are stitched together on-device into the final video output.
\end{enumerate}

\subsection{Preprocessing for Mobile Constraints}

To reduce inference latency and memory overhead, we preprocess video clips using the following strategies:

\begin{itemize}
\item \textbf{Resolution and frame rate reduction.} We downsample smartphone video input in both spatial and temporal dimensions.
\item \textbf{Clip formatting.} Each video clip is reshaped to a tensor of size $(1, 3, 6, 224, 224)$, corresponding to batch size, channels, sampled frames, height, and width. This size achieves a good trade-off between accuracy and efficiency for scene-level understanding.
\end{itemize}

The entire pipeline, from segmentation to script generation, runs fully on-device, benefiting from persistent module reuse to avoid unnecessary reloading. Figure~\ref{fig:video_assembly_pipeline} in the main paper visualizes this process.

\section{Broader Applicability to Video-Language Tasks}
\label{adx:broader_applications}

While video assembly is our primary case study, the reuse-centric architecture of \projectname generalizes naturally to a wide range of video-language tasks, including video retrieval, summarization, question answering, and more. 
These applications share a common multi-stage structure: they require video encoding followed by language-based reasoning or matching, making them well-suited for \projectname's modular, persistent, and parallel execution design.

In video retrieval, each clip is first processed by the visual encoder to produce embeddings. 
Instead of storing high-dimensional latent representations (e.g., 2048$\times$1024, which can occupy up to 8MB per clip), \projectname uses the language decoder to generate compact, semantically rich textual descriptions. 
This text-based representation reduces storage overhead to just a few hundred bytes per clip, and facilitates fast similarity-based matching using a lightweight sentence transformer model (e.g., \textsf{all-MiniLM-L6-v2}, $\sim$90MB). 
The reuse-centric pipeline reuses both the visual encoder and the language decoder across all retrieval steps, eliminating redundant model loading and significantly reducing latency and memory usage. 
This makes accurate video retrieval feasible even on memory-constrained mobile devices.

In video summarization, the model first generates scene-level captions for short video segments using the encoder-decoder pair.
These intermediate captions are then passed back into the same language decoder to generate higher-level summaries. 
Because the decoder remains loaded in memory from the initial captioning step, the summarization stage incurs no additional model load, leading to improved efficiency and consistent semantic representations across stages.
This reuse not only reduces overhead but also supports coherent output quality through the use of a single, high-capacity decoder across all stages.

For video-based question answering (VideoQA), the video is again segmented and captioned using the visual encoder and decoder. 
Given a user question, the decoder then performs reasoning based on either the generated captions or raw visual embeddings to generate an answer. 
Since the decoder is already resident from earlier stages, it can be directly reused for reasoning without loading any QA-specific models, thereby lowering both latency and memory consumption.

Across all these applications, the key advantage of \projectname lies in its system-level reuse strategy. 
By maintaining loaded modules persistently in memory, \projectname avoids the need to load and unload models at each subtask boundary. 
This leads to consistent reductions in runtime and memory footprint while simplifying the software pipeline. 
Furthermore, by reusing the same language decoder across tasks; captioning, summarization, retrieval, and reasoning,\projectname ensures semantic consistency across outputs and avoids divergence that could arise from using separate specialized models.

The generality of this design supports a broader vision: 
modular and unified execution for a spectrum of video-language understanding tasks, optimized for on-device deployment. 
This makes \projectname not just a mobile-friendly architecture for video assembly, but a reusable backbone for future multimodal applications across edge devices.

\section{Quantization}
\label{adx:quantization}

To enable deployment on commodity mobile hardware, we apply dynamic quantization to the video-language model components used in \projectname. 
Unlike previous work that focuses solely on shrinking model size for memory fit, our goal with quantization is to preserve runtime efficiency while maintaining compatibility with mobile backends. 
This ensures that reused modules can remain persistently loaded and operate efficiently across tasks.

Quantizing video-language models presents several unique challenges. 
First, many popular quantization methods lack support for key operators used in video encoders, such as \textsc{Conv3D}. 
Additionally, some libraries fail to support int8 quantization or lack serialization capabilities necessary for ARM-based mobile deployment. 
Finally, quantization must be performed in a way that minimally impacts model accuracy.

To address these challenges, we adopt a dynamic quantization strategy using \textsc{TorchAO}~\cite{torchao}. 
We target only the \textsc{Linear} and \textsc{Embedding} layers, which together account for over 91\% of model memory consumption. 
\textsc{TorchAO} supports these operators and provides compatibility with ARM CPU execution through int8 datatypes. 
It also supports serialization, allowing us to package and deploy quantized models across a range of Android devices.

We deliberately exclude quantization of the \textsc{Conv3D} layers due to unreliable support across quantization libraries. 
Instead, we address the computational cost of video encoding through preprocessing, including frame rate and resolution downsampling prior to inference. 
This approach reduces runtime load without sacrificing too much fidelity in scene understanding.

In our evaluation, quantization introduces only a modest performance drop, approximately 1.5 CIDEr points on the captioning task and 2.3 Recall@1 on video retrieval, while enabling full pipeline deployment on devices with just 6GB of RAM. 
We also observe faster model load times and reduced peak memory usage, which further improve end-to-end latency.

It is worth emphasizing that quantization in \projectname complements our core design goal: 
reducing runtime overhead through architectural reuse. 
Our contribution is not in developing new quantization methods, but in ensuring that standard techniques like dynamic quantization can be integrated seamlessly into a reuse-centric execution model, enabling persistent in-memory operation of high-capacity modules under mobile constraints.

\section{Serializing and Deploying the Quantized Model}
\label{adx:serialization}

On resource-constrained devices such as mobile phones, storing and deploying quantized models differs significantly from Python-based execution on high-end servers, where model weights are typically stored as a dictionary. 
For mobile deployment, the quantized model must be serialized by tracing its entire forward pass, including nested forward passes across multiple modules. 
This approach ensures compatibility with mobile execution environments and supports efficient inference on constrained hardware.

To enable saving and loading of quantized video-language models, we employed \texttt{torch.jit.trace}, which records the operations executed during a forward pass into a computational graph. However, several refactors were required to ensure compatibility:
\begin{enumerate}[leftmargin=0.5cm]
    \item \textbf{Statically-typed Forward Passes}: The PyTorch JIT (just-in-time) compiler struggles with dynamic datatype inference during variable assignments and conditional operations. We refactored all forward passes in the submodules (visual encoder, and text decoder) to use static typing.
    \item \textbf{Data-dependent Control Flows}: Tracing the text decoder proved challenging due to its inability to handle dynamic output lengths. As a workaround, we used sample inputs generating the maximum caption length and trimmed any excess tokens from the output.
    \item \textbf{Substitute Torch Operations}: Certain matrix operations, such as \texttt{torch.rsub} and \texttt{torch.zeros}, are incompatible with JIT tracing. We replaced these with equivalent operations, such as combining \texttt{torch.mul} and \texttt{torch.sub}, or using \texttt{torch.zeros\_like}.
\end{enumerate}
Following the tracing of the forward pass, we applied additional optimizations tailored for mobile deployments. These include: (a)~\textbf{Operator Fusion}: We merged the kernels of frequently paired operations, such as \textsc{Linear} and \textsc{ReLU}, to reduce execution overhead and enhance runtime efficiency.
(b)~\textbf{Parameter Hoisting}: To minimize the storage of structural metadata, we restructured the model by hoisting all operators to the top level of the forward pass, streamlining execution and reducing the overall memory footprint.
These optimizations further improve the efficiency and feasibility of deploying quantized video-language models on resource-constrained mobile devices.

\section{Qualitative Analysis}
\label{adx:qualitative-analysis}
\begin{figure}[!h]
    \centering
    \begin{subfigure}[t]{0.49\textwidth}
    \centering
    \includegraphics[width=1\linewidth]{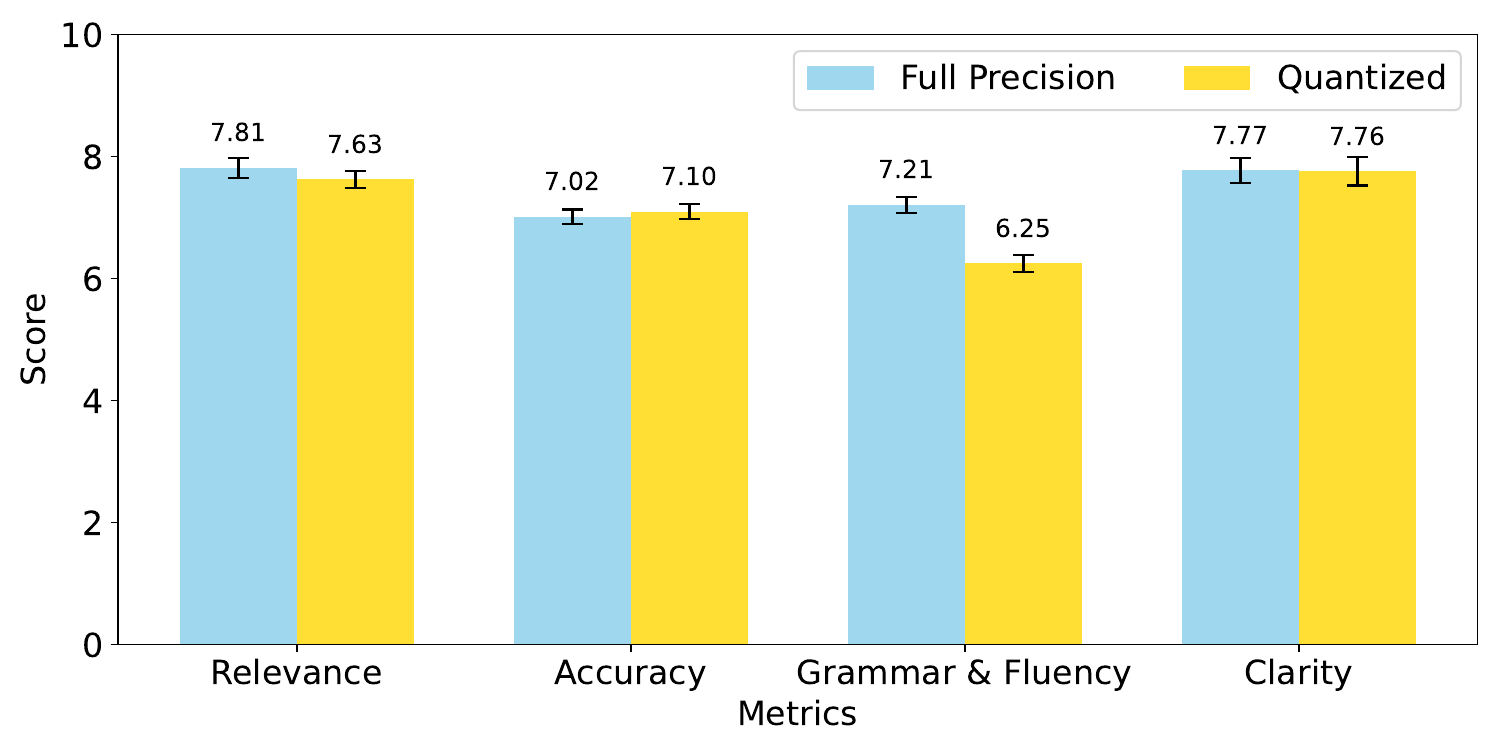}
    \caption{Video captioning}
    \label{subfig:qualitative-video-captioning}
    \end{subfigure}
    \begin{subfigure}[t]{0.49\textwidth}
        \centering
        \includegraphics[width=1\linewidth]{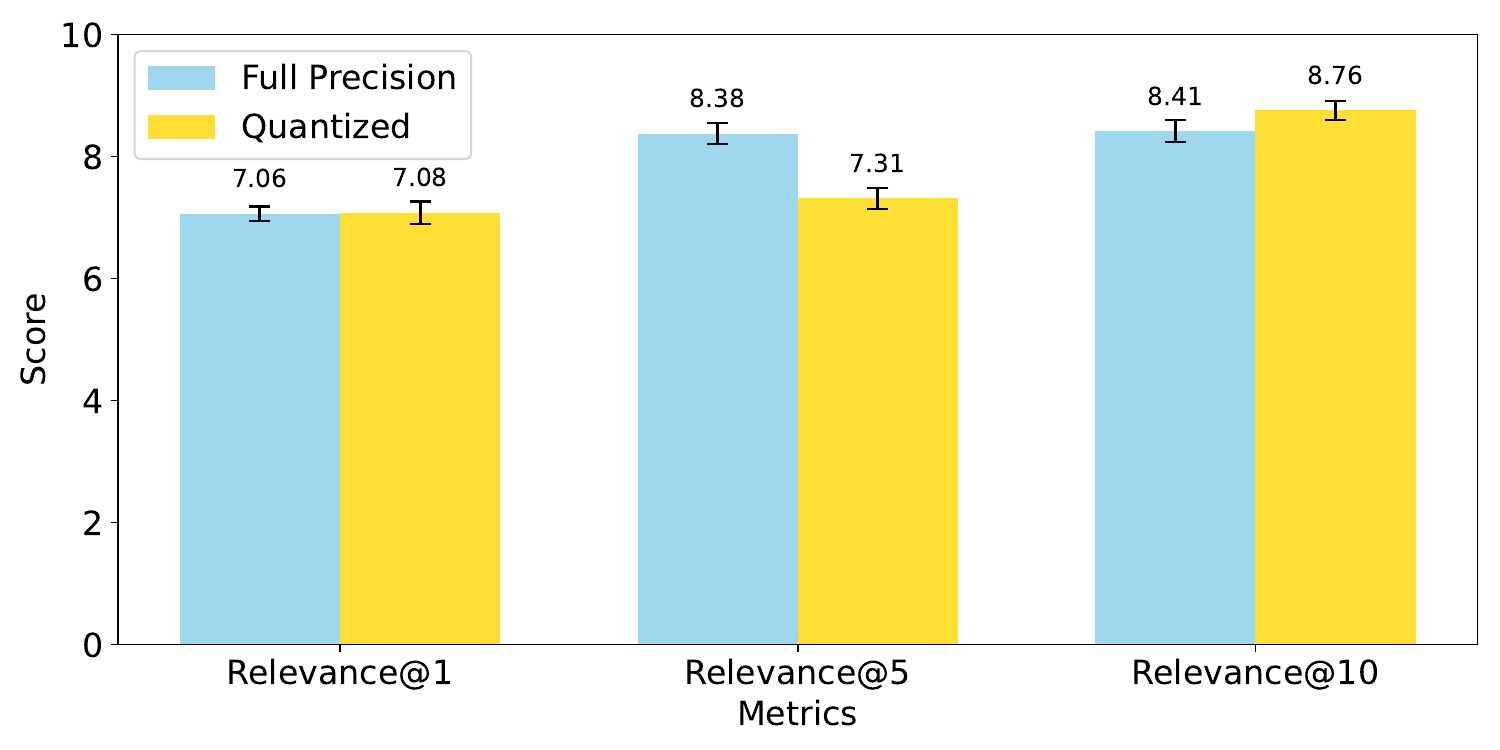}
        \caption{Video retrieval}
        \label{subfig:qualitative-video-retrieval}
    \end{subfigure}
    \begin{subfigure}[t]{0.49\textwidth}
        \centering
        \includegraphics[width=\linewidth]{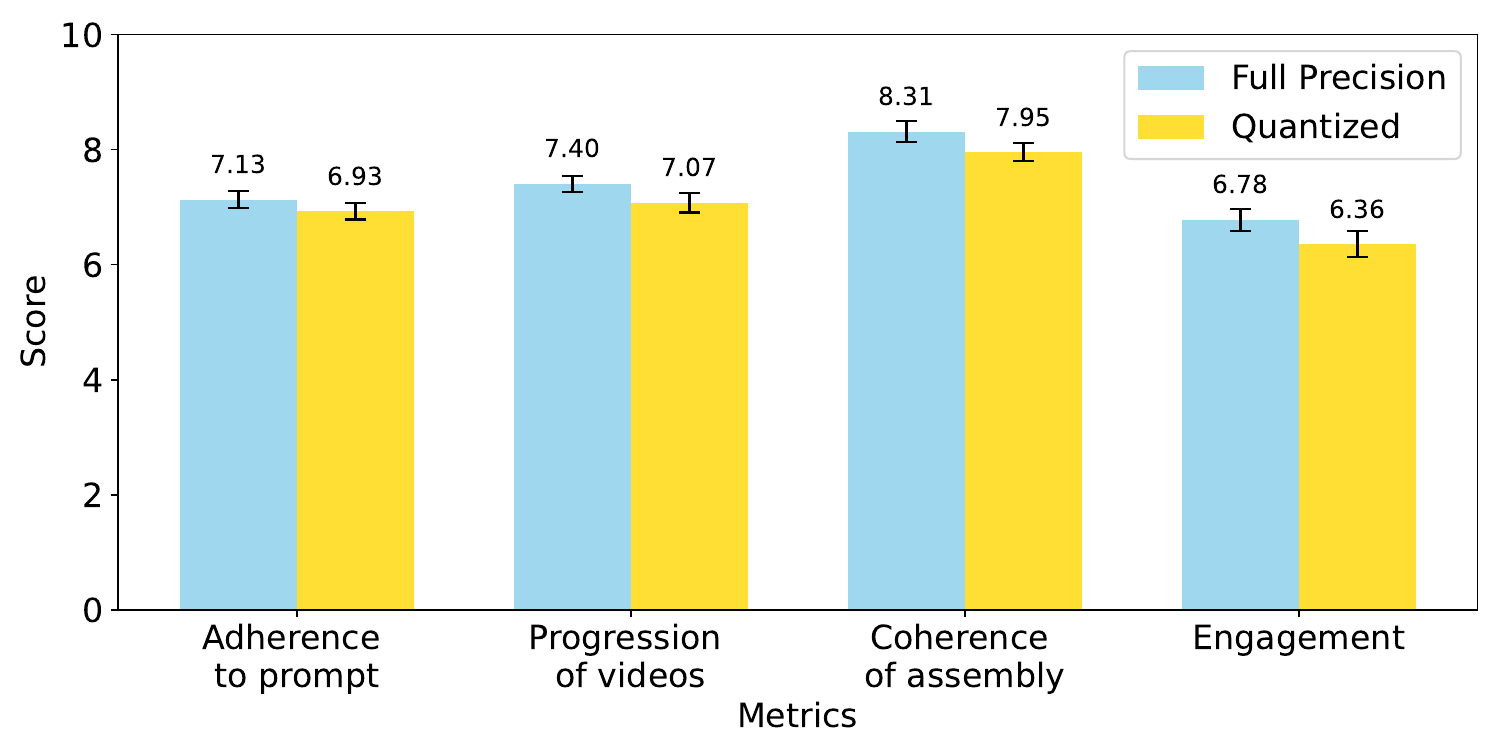}
        \caption{Video assembly}
        \label{subfig:qualitative-video-assembly}
    \end{subfigure}
    \caption{Qualitative analysis of the three tasks: (a) video captioning, (b) video retrieval, (c) video assembly. The results are based on 10 participants. The scale of different metric scores is [0, 10].}
    \label{fig:qualitative-analysis}
\end{figure}

With 10 human participants, we have evaluated the performance of video captioning, video retrieval, and video assembly tasks on the full-precision and quantized versions of \projectname's modified \textsf{mPLUG2+}.
The results are shown in Figure~\ref{fig:qualitative-analysis}.
The qualitative metrics have scores from 0 (low score for the corresponding metric) to 10 (high score for the corresponding metric).
The observed results related to the scores are discussed next. 

For the task of video captioning, we evaluated the generated captions on 
(a) Relevance: whether the caption in appropriate for the theme or essence of the video, 
(b) Accuracy: whether the caption has correct details about the objects present in the video, 
(c) Grammar and fluency: the linguistic correctness and naturalness of the caption, and 
(d) Clarity: whether the caption is unambiguous. Each participant was shown 10 videos with the captions generated from both the full-precision and quantized models. 
Figure~\ref{subfig:qualitative-video-captioning} presents the mean scores for the evaluated metrics, with error bars indicating standard deviation. 
The mean participant-assigned scores for full-precision and quantized models differ by 0.1-1 across the four metrics. 
The standard deviation of these scores ranges from 0.7 to 1.2 for both models, suggesting that the observed mean differences are statistically insignificant. 
Consequently, the results indicate a negligible difference in caption quality between full-precision and quantized models; echoing the quantitative results shown in Table 1.
Small samples of the generated captions are shown in Table~\ref{tab:qualitative-cpations}, corroborating the negligible differences between the generation capabilities of the full-precision and quantized models.

In context of the task of video retrieval, we evaluated the retrieved videos (given a prompt) on Relevance at top-$k$ (Relevance@$k$) by showing them top-$k$ video(s) and asking how appropriate the video content is to the written prompt. 
Each participant was asked to write 10 prompts and evaluate the top-1, top-5, and top-10 relevance on the videos retrieved by both the full-precision and quantized models. 
Figure~\ref{subfig:qualitative-video-retrieval} shows the mean scores for the above metrics, with error bars showing standard deviation.
The mean top-$k$ scores across all participants are in the range of 0.10-1.52 between full-precision and quantized models.
The standard deviation is in the range of 0.75-1.02 for the full-precision model, and in the range of 0.92-1.13 for the quantized model.
The results indicate that the mean top-$k$ scores for full-precision and quantized models fall within a similar range, with overlapping standard deviations.
This suggests that quantization has minimal impact on video retrieval performance, as both models exhibit comparable score distributions.

Lastly, for the task of video assembly, the qualitative evaluation is based on (a) Adherence to the prompt: Whether the content of the assembled video matched the given prompt, 
(b) Progression of videos: Whether the sequence of videos in the assembly seems to be in a natural order, 
(c) Coherence of the assembly: Whether the style and content of the assembled video clips make sense together, and 
(d) Engagement: How interesting or compelling the assembled video is.
Each participant was asked to write 10 prompts and evaluate the above metrics on the videos assembled by both the full-precision and quantized models. 
Figure~\ref{subfig:qualitative-video-assembly} shows the mean scores for the above metrics, with error bars showing standard deviation.
The mean scores for the full-precision model range from 6.8 to 8.3, while the quantized model scores range from 6.3 to 7.9. 
The standard deviation is between 0.81-1.17 for the full-precision model and between 0.77-1.19 for the quantized model. 
The overlapping standard deviations suggest that quantization has a negligible impact on the task of video assembly too, as both models yield similar score distributions.

\begin{table*}[h]
\centering
\small
\begin{tabular}{l|ll}
\toprule
\# & \begin{tabular}[c]{@{}l@{}}Caption generated by the original\\full-precision (float32) model\end{tabular} & \begin{tabular}[c]{@{}l@{}}Caption generated by the quantized\\quantized (int8) model\end{tabular} \\
\midrule
1 & a person is wrapping gifts &  a person is wrapping gifts \\
2 & showing a cityscape to the camera & a city is being shown \\
3 & a person is making a tortilla & a person is making a tortilla \\
4 & a woman unpacks a box & a woman opens a box  \\
5 & a woman is opening a box & a person is opening a box \\
6 & a woman is praying & a woman is putting candles on a statue \\
7 & a girl is reading a book & a girl is reading a book \\
8 & a woman is reading a book &  a woman is reading \\
9 & showing a photo of street fair to the camera & footage of a carnival \\
10 & turning the camera upwards while filming a city &  turning the camera upwards while filming a city \\
11 & a group of people are playing instruments outside & a group of people are singing and playing instruments \\
12 & a person is cooking & a person is cooking \\
\bottomrule
\end{tabular}
\caption{Comparison of the captions (text descriptions of the videos) generated by the original full-precision float32 \textsf{mPLUG} model and the quantized int8 video-language model (modified \textsf{mPLUG2+}). 
We see minute but synonymous differences, indicating that the quality of the captions are preserved after quantization.}
\label{tab:qualitative-cpations}
\end{table*}

Therefore, across all three tasks -- video captioning, video retrieval, and video assembly -- the qualitative evaluation shows that quantization has minimal impact on performance, with mean scores and standard deviations exhibiting significant overlap between full-precision and quantized models.

\section{Additional Results}
\label{adx:additional-results}

In this section, we present additional results that provide deeper insight into our system’s efficiency and design choices, including ablation studies, storage requirements, quantization effects, and memory footprint analysis.

\subsection{Ablation Studies}
\label{subsec:ablation-studies}
\begin{table}[!h]
\centering
\begin{tabular}{lll}
\toprule
\begin{tabular}[c]{@{}l@{}}Input Shape\\ (Batch size, Channels, FPS,\\\quad Width, Height)\end{tabular} & \begin{tabular}[c]{@{}l@{}}Latency\end{tabular} & \begin{tabular}[c]{@{}l@{}}CIDEr\\ Score\end{tabular} \\
\midrule
(1, 3, 6, 224, 224): default case & 5.57 & 82.5 \\
\midrule
(1, 3, 2, 224, 224) & 4.78 & 50.89 \\
(1, 3, 6, 112, 112) & 3.29 & 75.64 \\
(4, 3, 6, 224, 224) & 22.35 & 82.5 \\
\begin{tabular}[c]{@{}l@{}}(1, 3, 6, 224, 224):\\\quad generator beam size = 1\end{tabular} & 3.07 & 40.98 \\
\begin{tabular}[c]{@{}l@{}}(1, 3, 6, 224, 224):\\\quad traced output length = 12\end{tabular} & 4.55 & 74.19 \\
\bottomrule
\end{tabular}
\caption{Changing input sizes and hyperparameters to measure their impact on latency and performance. 
All experiments are measured on Google Pixel 5a for the task of video assembly with the test dataset of \textsf{MSR-VTT}.
The default input shape is (1, 3, 6, 224, 224), with traced output length of 23 and generated beam size of 4.}
\label{tab:ablation-studies}
\end{table}
Table~\ref{tab:ablation-studies} demonstrates the impact of varying input configurations and hyperparameters on the latency and performance of \projectname's modified \textsf{mPLUG2+} on a Google Pixel 5a. 
The default configuration {(1, 3, 6, 224, 224)} achieves a latency of 5.75s and a CIDEr score of 82.5, providing a baseline for comparison.

(i)~\textbf{Adjusting the traced output length} to 12 reduces latency to 4.55s, but lowers the CIDEr score to 74.19, as shorter outputs often fail to capture the full detail of ground truth captions.
(ii)~\textbf{Lowering the input frame rate (FPS)} from 6 to 2 reduces latency to 4.78s, but leads to a significant drop in performance (CIDEr: 50.89) due to insufficient temporal coverage.
(iii)~\textbf{Reducing the frame resolution} to \texttt{(112, 112)} lowers latency to 3.29s, with only a mild performance loss (CIDEr: 75.64), highlighting an effective trade-off.
(iv)~\textbf{Reducing the beam size} to 1 yields a latency of 3.07s, but severely degrades performance (CIDEr: 40.98) due to limited search diversity during generation.
(v)~\textbf{Increasing the batch size} to 4 increases latency to 22.35s, while maintaining the same CIDEr score (82.5) as the default, illustrating that batching offers no speed advantage on constrained hardware.
These results emphasize the importance of tuning both inputs and decoding settings to balance latency and quality, particularly in mobile deployment scenarios.

\subsection{Storage Size}
\label{subsec:storage-size}
\begin{table}[!h]
\small
\centering
\begin{tabular}{ll}
\toprule
 & \begin{tabular}[c]{@{}l@{}}Storage Size \\ (in MBs)\end{tabular} \\
 \midrule
Video-Language Model (\textsf{mPLUG+}) & 1230 \\
Sentence Transformer (\textsf{all-MiniLM-L6-v2}) & 53 \\
\midrule
Total & 1283 \\
\bottomrule
\end{tabular}
\caption{\textbf{Storage size (in MBs)} of the video-language model and sentence transformer used for end-to-end task pipelines of \textbf{video assembly}, \textbf{video captioning} and \textbf{video retrieval}. The total storage size required for the pipelines is 1.28GB, which is feasible for the most commodity smartphones.}
\label{tab:storage-size}
\end{table}
Table~\ref{tab:storage-size} shows that the combined storage size of the video-language model \textsf{mPLUG+} and sentence transformer \textsf{all-MiniLM-L6-v2} is 1.28GB, with the video-language model accounting for 1.23GB and the sentence transformer for 53MB. 
This compact size ensures that the models are feasible to store on most commodity smartphones, enabling on-device deployment of end-to-end pipelines for video captioning and retrieval.

\subsection{Quantizing the Sentence Transformer}
\label{subsec:sentence-transformer}
Table~\ref{tab:semantic-similarity-performance} shows that the full-precision version of \textsf{all-MiniLM-L6-v2} SentenceTransformer has a Pearson Correlation of 0.83 on the test partition of Semantic Textual Similarity Benchmark (\textsf{STS-B}) dataset.
The quantized version minutely underperforms with the Pearson Correlation of 0.81, however the qualitative analysis 
(in Appendix~\ref{adx:qualitative-analysis})
shows that the impact of such performance drops due to quantization is negligible on the end-to-end performance. 
\begin{table}
\centering
\begin{tabular}{ll}
\toprule
\begin{tabular}[c]{@{}l@{}}SentenceTransformer \\Variant\end{tabular} & \begin{tabular}[c]{@{}l@{}}Pearson \\Correlation ($\uparrow$)\end{tabular} \\
\midrule
Full-precision & 0.8287 \\
Quantized & 0.8156 \\
\bottomrule
\end{tabular}
\caption{Performance comparison of the full-precision and quantized versions of the sentence transformer \textsf{all-MiniLM-L6-v2} on Semantic Textual Similarity Benchmark (STS-B) dataset. 
The sentence transformer is used in our mobile pipelines to find text representation of the most relevant videos to a user query. We see a minor performance drop with the quantized version. 
While \projectname can accommodate the full-precision model in a mobile pipeline, we observe no end-to-end performance drop with the quantized sentence transformer.}
\label{tab:semantic-similarity-performance}
\end{table}

\subsection{Memory Footprint}
\label{adx:memory-footprint}
\begin{table}
\small
\centering
\addtolength{\tabcolsep}{-0.2em}
\begin{tabular}{lcc}
\toprule &
 \multicolumn{2}{c}{\begin{tabular}[c]{@{}c@{}}Max Memory\\ Consumption (in GBs)\end{tabular}} \\
 \cmidrule{2-3}
 & \begin{tabular}[c]{@{}c@{}} \projectname without \\ modularization \\ and reuse\end{tabular} & \begin{tabular}[c]{@{}l@{}} \projectname \end{tabular} \\
 \midrule
\begin{tabular}[c]{@{}l@{}}Model Weights\end{tabular} & 4.623 & 4.623 \\
Inputs & 0.005 & 0.005 \\
\begin{tabular}[c]{@{}l@{}}Activations\\\quad(highest across all the\\\quad models or modules)\end{tabular} & 0.297 & 0.479 \\
Miscellaneous\\\quad (temporary variables\\\quad in model loads) & 0.352 & 0.210 \\
\midrule
\begin{tabular}[c]{@{}l@{}}Peak Memory Consumption\\for One Run of the Pipeline\end{tabular} & 5.277 & 5.317 \\
\bottomrule
\end{tabular}
\caption{\textbf{Memory Consumption (in GBs, lower the better)} of the video-language model used for the task pipeline of \textbf{video assembly}. 
The pipeline with \projectname consumes 5.3GBs of RAM for \projectname, which is feasible to run for recent commodity smartphones having 6GB of RAM.
\projectname only incurs 0.76\% more memory usage compared to the implementation without modularization and reuse, due to the elimination of 40.34\% of the temporary allocations created during model loading.}
\label{tab:vlm-memory-consumption}
\end{table}

The goal of this section is to quantify the memory footprint of \projectname and show that its reuse-centric design does not exceed the tight RAM budgets of mobile devices.
Table \ref{tab:vlm-memory-consumption} reports the peak memory usage of the video-language pipeline with and without modular reuse. 
The two configurations differ by only 40MB: 5.28GB for the baseline versus 5.32GB for \projectname, a negligible 0.76\% increase that still fits comfortably within the 6 GB RAM of mainstream smartphones.
The extra 40MB are due to the activations being slightly higher under \projectname (0.479GB vs. the sequential baseline's 0.297GB) because the encoder and decoder can be active at the same time. 
This is offset by a sharp drop in the miscellaneous allocations, which are the temporary buffers created during model loading.
The max memory consumption related to the temporary allocations shrink by 40.34\% (\projectname's 0.210GB vs. the sequential baseline's 0.352GB), due to keeping the heavy modules permanently resident. 
Model weights and input memory remain identical across the two variants.

\paragraph{Latency–memory trade-off.}
Although \projectname sacrifices a 40MB extra RAM to enable parallelism, it delivers 27–33\% end-to-end latency reductions (see Section \ref{subsec:latency}), making the trade-off highly favorable. 
Tighter memory savings would require pushing quantization beyond 8-bit, which we found degrades accuracy unacceptably. 
Instead, \projectname prioritizes reuse to maximize speed while remaining within the practical 6GB envelope of commodity phones.

\subsection{Layerwise Memory Breakdown}
\label{subsec:layerwise-memory-breakdown}
This section provides a finer-grained analysis of memory consumption across different layers of the video-language model \textsf{mPLUG2+} used in \projectname.
This breakdown enabled the identification of the dominant contributors to the total memory footprint. 
Understanding this breakdown is essential for targeting further optimizations.

\begin{table}
\small
\setlength{\tabcolsep}{3.5pt}
\begin{tabular}{lccc}
\toprule
Module Name & \makecell{Parameter\\Count} & \makecell{Parameter Size\\(in MBs)} & \makecell{\% of Total\\Size} \\
\midrule
\textsc{Linear} & 1236M & 3689 & 80.82\% \\
\textsc{Embedding} & 161M & 487 & \phantom{}10.67\% \\
\textsc{Conv3d} & 51M & 194 & \phantom{0}4.25\% \\
\makecell{\textsc{\kern-1em NonDynamically} \\\textsc{QuantizableLinear}} & 49M & 193 & \phantom{0}4.22\% \\
\textsc{LayerNorm} & 0.4M & 2 & \phantom{0}0.04\% \\
\bottomrule
\end{tabular}
\caption{Layer-wise breakdown of peak memory consumption for \textsf{mPLUG2+}.}
\label{tab:layer-wise-breakdown}
\end{table} 

Table~\ref{tab:layer-wise-breakdown} shows that the \textsc{Linear} and \textsc{Embedding} layers dominate memory usage, consuming 80.82\% (3,689MB) and 10.67\% (487MB) of the total size, respectively. These layers, which are extensively used in the text decoder, present significant optimization opportunities.

By reusing the text decoder, \projectname effectively reduces both memory consumption and execution latency, demonstrating the efficiency of this architectural choice. This insight reinforces the importance of targeted optimizations in deep learning pipelines for resource-constrained environments.

\subsection{Extended Results: Video Retrieval on DiDeMo dataset.}
\label{adx:extended-results-didemo}
To complement the main table on the video retrieval performance (Table~\ref{tab:retrieval-performance}) in Section~\ref{subsec:performance}, Table~\ref{tab:retrieval-performance-extended} presents the full DiDeMo retrieval results, highlighting how \textsf{mPLUG2+} maintains strong performance even under quantization.

\begin{table}[h]
\centering
\footnotesize
\begin{tabular}{>{\raggedright\arraybackslash}p{2cm}  
  >{\centering\arraybackslash}p{1cm}    
  >{\centering\arraybackslash}p{1cm}    
  >{\centering\arraybackslash}p{1cm}
}
\toprule
\multirow{2}{*}{\makecell{Model\\Variant}} & \multicolumn{3}{c}{DiDeMo} \\ \cmidrule{2-4}
 & R@1 & R@5 & R@10 \\ \midrule
\multicolumn{4}{c}{Without Quantization (32-bit floating point weights)} \\ \midrule
\textsf{mPLUG2} & 55.1 & 78.3 & 84.6 \\
\textsf{mPLUG2+} & 60.0 & 86.2 & 90.1 \\ \midrule
\multicolumn{4}{c}{With Quantization (8-bit integer weights)} \\ \midrule
\textsf{mPLUG2} & 53.7 & 75.1 & 80.8 \\
\textsf{mPLUG2+} & 58.7 & 83.1 & 88.4 \\ \bottomrule
\end{tabular}
\caption{\textbf{Performance} comparison of the original \textsf{mPLUG2} (with \textsf{BERT} text decoder) and our modified \textsf{mPLUG2+} (with \textsf{Llama} text decoder) for both quantized and full-precision versions for the task of \textbf{video retrieval} on DiDeMo dataset. 
All 3 metrics R@1 (Recall@1), R@5 (Recall@5), and R@10 (Recall@10) are ``higher the better''.
\textsf{mPLUG2+} of \projectname has a lower performance drop even after quantization, compared to the original \textsf{mPLUG} with a weaker text decoder.
}
\label{tab:retrieval-performance-extended}
\end{table}

\subsection{Estimated Energy Consumption on Pixel 5a}
\label{adx:energy-consumption}

To quantify the energy efficiency benefits of \projectname on a commodity mobile device, here we compute the reduction in energy consumption during video-language processing on the Google Pixel 5a.

\begin{itemize}
    \item \textbf{CPU Power:} The Pixel 5a’s Snapdragon 765G CPU draws approximately 2 Watts under moderate computational load typical of video-language model inference.
    \item \textbf{DRAM Power:} Active memory power consumption for LPDDR4X RAM is roughly 0.2 Watts at 6GB utilization during active workloads.
    \item \textbf{Latency:} The original pipeline execution latency is \(T_{\text{orig}} = 10\) seconds (example value). \projectname reduces latency by 27–33\%, yielding a new latency \(T_{\text{new}} \approx 6.7\text{–}7.3\) seconds.
    \item \textbf{Memory Footprint:} \projectname’s peak RAM usage is essentially unchanged: 5.277GB (baseline) versus 5.317GB (+0.76\%). The 0.76\% increase is small enough that we assume DRAM power remains at 0.2W.
\end{itemize}

\paragraph{Calculation:}
\[
\begin{aligned}
\text{Original Energy} &= (P_{\text{CPU}} + P_{\text{DRAM}}) \times T_{\text{orig}} \\
&= (2.0\,\text{W} + 0.2\,\text{W}) \times 10\,\text{s} = 22\, \text{J} \\
\\
\text{Adjusted CPU Power} &= 2.0\,\text{W} \times (1 - 0.27) \approx 1.46\,\text{W} \\ \text{(assuming} &\text{ CPU power scales with latency)} \\
\text{Adjusted DRAM Power} &= 0.2\,\text{W} \times (1 - 0.057) \approx 0.19\,\text{W} \\
\\
\text{New Energy} &= (1.46\,\text{W} + 0.19\,\text{W}) \times 7.0\,\text{s} \\&= 11.9\, \text{J} \\
\\
\therefore \text{Energy Savings} &= \frac{22 - 11.9}{22} \times 100\% \approx 46\%
\end{aligned}
\]

This estimate assumes linear scaling of CPU power with latency reduction and memory power with memory footprint reduction, which is reasonable for short, compute-bound workloads.

\paragraph{Implications:}

Reducing energy consumption by nearly half per pipeline execution can significantly extend battery life during repeated video-language tasks.
Most of the savings stem from \projectname's reuse-centric design that trims CPU active time; memory power remains virtually unchanged because the peak footprint grows by only 0.76\%.

\section{Hyperparameter Selection.}
\label{adx:hyperparameter-selection}

In order to train \textsf{mPLUG2+} model on caption generation task, we had conducted a grid search over learning rates \{1e-3, 1e-4, 1e-5, \textbf{2e-5}, 5e-5, 1e-6\} and found 2e-5 to yield the best tradeoff between stability and performance. 
Training beyond 10 epochs led to overfitting, so we used the checkpoint at epoch \textbf{10}. 
Batch sizes of \{32, 64, \textbf{128}, 256\} were also evaluated; the batch size of 128 provided the best balance between training time and CIDEr score. 
The impact of other video-related hyperparameters is analyzed in the ablation study (Appendix~\ref{subsec:ablation-studies}).

\newpage
\bibliography{aaai2026}


\end{document}